\documentclass[11pt]{article} % DO NOT CHANGE THIS
\usepackage{fullpage}
\usepackage{libertine}  % DO NOT CHANGE THIS
\usepackage{graphicx} % DO NOT CHANGE THIS
\usepackage{url}
\usepackage{caption} % DO NOT CHANGE THIS AND DO NOT ADD ANY OPTIONS TO IT

\date{}
\usepackage{amsmath,amssymb,amsthm}
\usepackage[algoruled,vlined,nofillcomment]{algorithm2e}
\usepackage{algpseudocode}
\newtheorem{definition}{Definition}

\newcommand{\cW}{\mathcal{W}}
\newcommand{\R}{\mathbb{R}}

\renewcommand{\Pr}{\mathsf{Pr}}
\newcommand{\norm}[1]{\| #1 \|}
\newcommand{\argmin}{\mathop{argmin}}

\newcommand{\eg}{e.g.,}

\newcommand{\ie}{i.e.,}

\usepackage{acronym}
\acrodef{DP}{Differential Privacy}
\acrodef{LDP}{Local \ac{DP}}
\acrodef{mDP}{Metric \ac{DP}} % TD: I think MDP would actually look less jarring, but happy to stick with mDP
\acrodef{NLP}{Natural Language Processing}
\acrodef{ML}{Machine Learning}
\acrodef{AI}{Artificial Intelligence}
\acrodef{biLSTM}{Bidirectional LSTM} % TD: Are we happy with not spelling out LSTM?
\acrodef{GESD}{Geometric mean of Euclidean and Sigmoid Dot product}
\acrodef{MAP}{Mean Average Precision}
\acrodef{MRR}{Mean Reciprocal Rank}
\acrodef{PII}{\emph{personally identifiable information}}
\acrodef{AUC}{Area Under Curve}

\title{Research Challenges in Designing \\ Differentially Private Text Generation Mechanisms}
\author{Oluwaseyi Feyisetan, Abhinav Aggarwal,  Zekun Xu, Nathanael Teissier\\Amazon\\\{sey,aggabhin,zeku,natteis\}@amazon.com}
\begin{document}

\maketitle

\begin{abstract}
Accurately learning from user data while ensuring quantifiable privacy guarantees provides an opportunity to build better \ac{ML} models while maintaining user trust. Recent literature has demonstrated the applicability of a generalized form of Differential Privacy to provide guarantees over text queries. Such mechanisms add privacy preserving noise to vectorial representations of text in high dimension and return a text based projection of the noisy vectors. However, these mechanisms are sub-optimal in their trade-off between privacy and utility. This is due to factors such as a fixed global sensitivity which leads to too much noise added in dense spaces while simultaneously guaranteeing protection for sensitive outliers.
In this proposal paper, we describe some challenges in balancing the tradeoff between privacy and utility for these differentially private text mechanisms. At a high level, we provide two proposals: (1) a framework called \textsc{lac} which defers some of the noise to a privacy amplification step and (2), an additional suite of three different techniques for calibrating the noise based on the local region around a word. Our objective in this paper is not to evaluate a single solution but to further the conversation on these challenges and chart pathways for building better mechanisms.
\end{abstract}

\section{Introduction}
Privacy has emerged as a topic of strategic consequence across all computational fields -- from machine learning, to natural language processing and statistics. Whether it is to satisfy compliance regulations, or build trust among customers, there is a general consensus about the need to provide privacy guarantees to users whose datasets serve as inputs to arbitrary functions provided by external processors. Within the mathematical and statistical disciplines, Differential Privacy \cite{dwork2006calibrating} has emerged as a gold standard for evaluating theoretical privacy claims. At a high level, a randomized algorithm is differentially private if its output distribution is \emph{similar} when the algorithm runs on two neighboring input databases. The notion of similarity is controlled by a parameter $\varepsilon \geq 0$ that defines the strength of the privacy guarantee. Similarly, it is possible to train differentially private deep learning \cite{shokri2015privacy,abadi2016deep} models by extending the methods from the statistical literature to the universal function approximators in neural networks. However, while \ac{DP} comes with strong theoretical guarantees, and the literature around the field is quite mature, designing differentially private mechanisms for generating text is less studied. 

As a result, within the field of traditional and computational linguistics, the norm is to apply anonymization techniques such as $k$-anonymity \cite{sweeney2002k} and its variants. While this offers a more intuitive way of expressing privacy guarantees as a function of an aggregation parameter $k$, all such methods are provably non-private \cite{korolova2009releasing}. Nevertheless, recent works such as \cite{fernandes2018generalised,icdm_paper,wsdm_paper} have attempted to directly adapt the methods of \ac{DP} to \ac{NLP} by borrowing ideas from the privacy methods used for location data \cite{andres2013geo}. In \ac{DP}, one way privacy is attained by adding `properly calibrated noise' to the output of a mechanism \cite{dwork2006calibrating}, or to gradient computations for deep learning \cite{abadi2016deep}. The premise of such `\ac{DP} for text' methods is predicated on adding noise to the vector representation of words in a high dimensional embedding space, and projecting the noisy vectors back to the discrete vocabulary space. 

Unlike statistical queries however, language generation comes with a unique set of problems. Consider a simple counting query where the objective is to return the number of people who exhibit a certain property $x$. The sensitivity of such a query is $1$ since a new individual can only increase the count by $1$. With text however, the sensitivity is much larger and is driven by the richness of the vocabulary, and how it is represented in the metric space under consideration. In this paper we propose strategies for increasing the utility of these 'DP for text' mechanisms by reducing the noise required while maintaining the desired privacy guarantees.

\section{Privacy Implications and Threat Model}
%Our threat model can be scoped within one of the modular units in an \ac{ML} workflow as defined by \cite{kairouz2019advances}. 
We consider a system where users generate training data (as text) which is then made available to an analyst. The analyst's utility requirement is to assess the quality of a downstream metric (e.g., ML model accuracy) derived from this data. The analyst therefore requires clear form access to the input data (e.g., for augmentation, aggregation, or annotation) to continuously improve downstream utility. In this model however -- akin to a modular unit in the overall threat vector of \cite{kairouz2019advances} and specific model of \cite{wsdm_paper}, it is possible that the analyst learns more information about the user e.g., their \emph{identity}, or some \emph{property} \cite{wagner2018technical}, than is required to play their role of improving the utility metric. An example where textual data was used for re-identification can be seen with the AOL data release \cite{barbaro2006face}.

\section{Challenges in Designing Private Text}
Consider a set of $n$ users, each with data $x_i \in \mathcal{X}$. Each user wishes to release up to $m$ messages in a privacy preserving manner while maximizing the utility gained from the release of the messages. One approach is for each user to submit their messages $(x_i,_1, \ldots, x_i,_m)$ in clear form to a \emph{trusted} curator. The curator then proceeds to apply a privacy preserving randomized mechanism $\mathcal{R}(*)$ to the analysis $\mathcal{A}(x)$ on the aggregated data. The privacy mechanism works by \emph{injecting noise} to the results of the analysis. This technique corresponds to the curator model of \ac{DP} \cite{dwork2006calibrating}, however, it requires that the users \emph{trust} the curator. This is the proposed approach for preserving privacy in the upcoming U.S. Census \cite{abowd2018us}. The curator model results in high utility since noise is applied only once on the aggregated data; however, a parallel approach cannot be clearly drawn for private text synthesis.

Another theoretical approach is for each user to apply the encoding or randomizing mechanism $\mathcal{R} : \mathcal{X} \to \mathcal{Y}^m$ to their data. The resulting $n \cdot m$ messages $(y_i,_1, \ldots, y_i,_m) = \mathcal{R}(x_i)$ for each user is then passed to the curator for analysis $\mathcal{A} : \mathcal{Y}^* \to \mathcal{Z}$. This corresponds to \ac{LDP} \cite{kasiviswanathan2011can}, since each user randomizes their data \emph{locally}. The model provides stronger privacy guarantees in the presence of an \emph{untrusted} curator. However, it incurs more error than the curator model because it requires multiple local $\mathcal{R}(x_i)$ transformations (as opposed to one by the trusted curator).
As a result, it has mainly been successfully adopted by companies with large user bases (such as Microsoft \cite{ding2017collecting}, Google \cite{erlingsson2014rappor}, and Apple \cite{appleDP}) which compensate for the error. The local model is more amenable for text \cite{fernandes2018generalised} and the literature builds on this framework.

The error accrued in the local model is exacerbated by the output range of the randomization function $\mathcal{R}(x_i)$. As an example, for one-bit messages (\eg{} a coin flip) where $f : \mathcal{X} \to \{0, 1\}$,
%\abhi{Did you mean $f : \mathcal{X} \to \{0, 1\}$?}, 
the overall error goes down faster as the number of users increase, given the small output size of $2$. Using a die roll with $6$ outputs, the noise smooths out a bit slower. However, for analysis over vector representations of words $f : \mathcal{X} \to \mathbb{R}^d$, where $d$ is the dimensionality of a word embedding model, and the number of words in the vocabulary could exceed thousands, the resulting analysis leads to far more noisy outputs. The noise (and by extension, the error) increases because of the \ac{DP} promise, \ie{} to guarantee privacy and protect all outliers, there must be a non-zero probability for transforming any given $x$ to \emph{any} other $x'$. We loosely correlate this size of the output space with the \emph{sensitivity} of the function $f$.
% (see \Cref{def:global_sensitivity} for a formal definition). 
Therefore, when the sensitivity is large, more noise is required to preserve privacy.

The challenge with designing privacy mechanisms for text stems from these aforementioned issues. We observe that unlike the natural distribution of values over the number line, the vector representation of words in an embedding space tends to be non-uniform. The distance between words carries information as to their semantic similarities, and as a result, there are sparse regions and dense regions. Conversely, the privacy guarantees from differential privacy extends to every word in the entire space (leading to the large noise required to ascertain worst-case protections). This problem is not unique to the text space, however, it has been better studied in the statistical privacy literature. For example, the theoretical sensitivity for computing the median of an arbitrary set of numbers is infinite, but, in most dataset scenarios, the sensitivity is smaller as values coalesce around the median \cite{nissim2007smooth}. Similar considerations have also been explored in private release of graph statistics \cite{blocki2013differentially,kasiviswanathan2013analyzing}.

%One way to reduce the error is to limit the number of contributions $m$ from each of the $n$ users, thereby putting a bound on the amount of privacy preserving noise required. This approach was adopted by \cite{korolova2009releasing} for privately releasing the head of search logs, and in the context of neural networks by \cite{abadi2016deep} and \cite{geyer2017differentially}. However, capping user contributions introduces bias into the data (by equating power users with large submissions, to tail users with a few contributions), and this bias-variance trade-off needs to be handled separately \cite{amin2019bounding}. In this work, we assume that all users submit the same number of data points, and we are primarily concerned with the noise associated with the \emph{sensitivity} over each datum (and not the number of contributions). Therefore, another way to reduce the error is to simply reduce the magnitude of the noise: this \emph{theoretically} leads to weakened privacy guarantees, however, it could be a practical solution for some empirical cases where the \emph{typical} sensitivity is less than the theoretical sensitivity \cite{amin2019bounding}. For example, the theoretical sensitivity for computing the median of an arbitrary set of numbers is infinite, but, in most dataset scenarios, the sensitivity is smaller as values coalesce around the median \cite{nissim2007smooth}. %We now discuss some related work on achieving privacy while limiting the magnitude of injected noise.

In this exploratory paper we examine these challenges from different lenses: 
\begin{enumerate}
\item Can we reduce the noise by deferring additional privacy guarantees to other amplification mechanisms that do not require noise (e.g., sub-sampling, shuffling, k-aggregation have all been proposed in the literature \cite{li2012sampling,bittau2017prochlo});
\item Can we re-calibrate the noise added such that it varies for every word depending on the density of the space surrounding the current word -- rather than resorting to a single global sensitivity?
\end{enumerate}

To address (1), we propose framing the private data release problem within the central \ac{DP} \cite{erlingsson2019amplification} paradigm by recommending a generalized form of the ESA protocol of \cite{bittau2017prochlo} which we denote as LAC. For (2), we propose three different methods that can be adopted to directly reduce the noise: density modulated noise, calibrating the noise to data sensitivity, and truncating the noise using a variety of approaches. 

We now give some preliminaries before expounding into the details of our proposals.

\section{Preliminaries and Current Methods}
%Adjacent DB
%Differential Privacy
%Local Model
%Curator Model
%Shuffled Model
%Generalised Metric Space
%L1 sensitivity
%Metric L1 sensitivity
%Smooth sensitivity

Let $\mathcal{X}^n$ be a collection of datasets from users. The Hamming distance $d_H(x,x')$ between two datasets $x,x' \in \mathcal{X}^n$ is the number of entries on which $x$ and $x'$ differ \ie{} $d_H(x,x') = | \{i : x_i \neq x'_i \} | = \sum_{i=1}^{|\mathcal{X}|} |x_i -x_i'|$. The datasets $x, x' \in \mathcal{X}$ are adjacent (we denote this as $x \sim x'$) if $d_H(x,x') = 1$.

\begin{definition}[Differential Privacy \cite{dwork2006calibrating}]\label{defn:dp}
A randomized algorithm $\mathcal{A} : \mathcal{X}^n \to \mathcal{Z}$ is $\varepsilon$-differentially private if for every pair of adjacent datasets $x \sim x' \in \mathcal{X}^n$ and every $\mathcal{Z} \subseteq Range(\mathcal{A})$, %it holds that
\begin{align*}
\Pr[\mathcal{A}(x) \in \mathcal{Z}] \leq e^\varepsilon \Pr[\mathcal{A}(x') \in \mathcal{Z}] .
\end{align*}
\end{definition}

\noindent A \ac{DP} algorithm protects a user by ensuring that its output distribution is approximately the same, whether or not the user was in the dataset used as an input to the algorithm. \ac{DP} is usually achieved by applying noise drawn from a Laplace distribution scaled by the sensitivity of the analysis function.

Several pieces of research have demonstrated generalized \ac{DP}  (also known as $d_{\mathcal{X}}$ privacy) for different metric spaces and distance functions \cite{chatzikokolakis2013broadening,andres2013geo,chatzikokolakis2015constructing,wsdm_paper,fernandes2018generalised,icdm_paper}. For example, \cite{chatzikokolakis2013broadening} demonstrated how the Manhattan distance metric was used to preserve privacy when releasing the number of days from a reference point. Similarly, the Chebyshev metric (chessboard distance) was adapted to perturb the output of smart meter readings \cite{wagner2018technical,chatzikokolakis2013broadening} providing privacy with respect to TV channels being viewed. Further, the Euclidean distance was utilized by \cite{andres2013geo,chatzikokolakis2015constructing} in a $2$ dimensional coordinate system to privately report the location of users, and finally, \cite{fernandes2018generalised} applied the Wasserstein metric in higher dimensions to demonstrate privacy preserving textual analysis using the metric space realized by word embeddings.

This work focuses on preserving privacy in high dimensional metric spaces equipped with the Euclidean metric. To achieve this form of metric differential privacy ($d_{\mathcal{X}}$ privacy), using a corollary to the Laplace mechanism, noise is sampled from an $n-$dimensional Laplacian and added to the output of the desired mechanism. %The noise is controlled by the privacy parameter $\varepsilon$ and the metric sensitivity $\Delta_{\mathcal{M}_f}$ \abhi{This is undefined.}.

%\begin{definition}[Multivariate Laplace mechanism \cite{andres2013geo}]
%Given the $n-$dimensional Laplacian $Lap^n_{\varepsilon}(x|\lambda)$ realized from a vector in the unit hypersphere $\mathbb{B}^n$ and magnitude sampled from the Gamma distribution $Gam^n_{1/\varepsilon}(x)$, the algorithm $\mathcal{A}(\mathcal{X}) = f(\mathcal{X}) + \eta$ is $d_{\mathcal{X}}$ private. Where the privacy preserving noise $\eta \sim Lap^n_{\varepsilon}(x|\lambda)$
%\end{definition}
%
%With these definitions, we now present an overview to our composite $d_{\mathcal{X}}$-privacy mechanism.

\section{Proposal 1: Deferred Amplification}

Our mechanism starts with a protocol similar to the privacy strategy of the local model. Given a set of $n$ users, each with $m$ data submissions $x_i \in \mathcal{X}$. Each user applies the $d_{\mathcal{X}}$ privacy mechanism $\mathcal{L} : \mathcal{X} \to \mathcal{Y}^m$ to their data. The resulting $n \cdot m$ messages $(y_i,_1, \ldots, y_i,_m) = \mathcal{L}(x_i)$ for each user is then passed to the curator $\mathcal{C} : \mathcal{Y}^* \to \mathcal{Z}$. 

Our proposal includes an additional step that amplifies the privacy guarantees. Between the local noise injection $\mathcal{L}(x)$ and the curator analysis $\mathcal{C}(y)$, we introduce a privacy amplification step $\mathcal{A} : \mathcal{Y}^* \to \mathcal{Y}^*$ which takes in the result of the message perturbations from all the users $\mathcal{A}(\cup^n_{i=1} \mathcal{L}(x_i))$, amplifies the privacy and outputs it to the curator. 

  \begin{figure}[t]
  \centering
    \includegraphics[width=0.5\columnwidth]{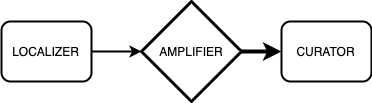}
    \caption{The \textsc{lac} privacy mechanism consists of a local randomizer which adds the noise, a privacy amplification module and a curator that aggregates the data.}
    \label{fig:composite_madlib}
  \end{figure}

To get an intuition on how \textsc{lac} can be used to improve utility while preserve privacy, consider the standard randomized response of \cite{warner1965randomized}. Given a bit $b \in \{0, 1\}$ and privacy parameter $\varepsilon$. To output a privatized bit $\hat{b}$, we set $\hat{b} = b$ with probability $p = \frac{e^{\varepsilon}}{1 + e^{\varepsilon}}$, otherwise $\hat{b} = 1 - b$. To improve the utility of this mechanism, we need to increase $\varepsilon$. However, in the local model, an adversary can map the output $\{\hat{b_1}, \ldots, \hat{b_n}\}$ to the $n$ corresponding users. Therefore, the parameter $p$ has to be close to $\frac{1}{2}$ otherwise $\hat{b} \approx b$ and the privacy guarantees are meaningless.
 Thus, to maintain the original (privacy) guarantees (while improving the utility), we need an additional mechanism that's different from the bit flipping noise addition. The desired property is such that the privacy guarantees are still meaningful when $p \ll \frac{1}{2}$. 
 
In building composite \ac{DP} algorithms, tools for \emph{privacy amplification} are used to design mechanisms that provide additional guarantees than the initial privacy protocol.

% \begin{itemize}
% \item Sub-sampling \cite{chaudhuri2006random,kasiviswanathan2011can,balle2018privacy}
% \item Iteration \cite{feldman2018privacy}
% \item Shuffling \cite{bittau2017prochlo,erlingsson2019amplification,cheu2019distributed,balle2019privacy}
% \end{itemize}

Probably the most studied technique is privacy `amplification by sub-sampling' \cite{chaudhuri2006random,kasiviswanathan2011can,balle2018privacy}, which states in its basic form that an $\varepsilon$-\ac{DP} mechanism applied to a $q$ fraction sub-sample of the initial population, yields an $\varepsilon'$-\ac{DP} mechanism, where $\varepsilon' \approx q\varepsilon$. Other approaches such as \cite{li2012sampling} and \cite{feyisetan2019privacy} have proposed augmenting sub-sampling with a $k-$anonymity parameter. Another class of amplification is by \emph{contractive iteration} \cite{feldman2018privacy} for privacy preserving \ac{ML} models. 

%   \begin{figure*}[h!]
%   \centering
%     \includegraphics[width=0.9\textwidth]{images/shuffle}
%     \caption{The privacy mechanism of Alg~\ref{alg:composite_madlib} consists of a local randomizer which adds the DP noise, a shuffler for privacy amplification and a curator that aggregates the data.}
%     \label{fig:composite_madlib}
%   \end{figure*}

\begin{algorithm}[t]
\footnotesize
\DontPrintSemicolon
\SetKw{KwRelease}{release}
\caption{Composite privacy mechanism}\label{alg:composite_madlib}
\tcp{Localizer}

\KwIn{word $w \in \mathcal{W}$, parameters $m$, for each $n$ users}
\KwOut{word $\hat{w} \in \mathcal{W}$}
\For{$i \in \{1, \ldots, m\}$}{
Noise $\eta \sim \text{Lap}(\Delta_{f}/\varepsilon)$\; 
$\hat{\phi} = \phi(w) + \eta$\;
}
\KwRelease{$\hat{w}$}
\;
\;
\tcp{Amplifier}
\KwIn{Multiset $\{\hat{w_i}\}_{i \in [n]}$, outputs of randomizers}
\KwOut{Multiset $\{\hat{w_i}\}_{i \in [n]}$, uniform permutes of [n]}
\For{$i \in \{n - 1, \ldots, 1\}$}{
$j \leftarrow$ random integer such that $0 \leq j \leq i$ \;
exchange $w_i$ and $w_j$\;
}
\KwRelease{$\{w\}$}
\;
\;
\tcp{Curator}
\KwIn{Multiset $\{y_i\}_{i \in [n]}$, with $y_i \in \mathcal{Y}$}
Compute $z = \mathcal{A}(y)$\;
\KwRelease{$z$}
\end{algorithm}

\subsection{Amplification Model Spotlight: The Shuffler}\label{sec:privacy_amplification}
In this work, we highlight the \emph{shuffle} mechanism \cite{bittau2017prochlo,erlingsson2019amplification,cheu2019distributed,balle2019privacy} to amplify the privacy guarantees. While shuffling on its own offers no \ac{DP} guarantees (unlike sub-sampling, which does), when combined with \ac{LDP}, it has the advantage of maintaining the underlying statistics of the dataset by not `throwing away' any of the data. The shuffler de-links data by masking its origin and confounding its provenance. For shuffling to be a viable amplification model, the Analyzer and Randomizer outputs must be amenable to shuffling, and not rely on any discriminating characteristics that link an individual to their contributions.

The pseudo-code in Alg.~\ref{alg:composite_madlib} provides a high level overview of the composite privacy mechanism using a shuffler. Each user contributes their data which passes through a local privacy randomizer. The noisy outputs are then passed to a shuffler which permutes the order of the source of the perturbed data. The overall protocol $\mathcal{P}$, thus, consists of ($\mathcal{L}$, $\mathcal{A}$, $\mathcal{C}$) and is modeled around the \emph{Encode, Shuffle, Analyze} (\textsc{esa}) architecture of \cite{bittau2017prochlo}.

% \begin{definition}[Amplification by shuffling \cite{erlingsson2019amplification}]
% For a domain $\mathcal{D}$, let $\mathcal{R}^{(i)} : \cW^{(1)} \times \ldots \times  \cW^{(i-1)} \times \mathcal{D} \to \cW^{(i)}$ for $i \in [n]$ be a sequence of $\varepsilon_0-$\ac{DP} algorithms. Let $\mathcal{S} : \mathcal{W}^* \to \mathcal{W}^*$ be an algorithm that samples a uniform permutation $\pi$ over [n]. Then algorithm $\mathcal{S}$ satisfies $(\varepsilon,\delta)$-\ac{DP} where
% %{\small
% \begin{align*}
% \varepsilon = \mathcal{O} \Bigg( \frac{\varepsilon_0 \sqrt{\log(1/\delta)}}{\sqrt{n}} \Bigg)
% \end{align*}
% %}
% \end{definition}

% \noindent and $n > 1, \delta > 0$ and $\delta$ is much smaller than $1/|D|$. The definition derives from the advanced composition \cite{dwork2010boosting} property of differential privacy for $k$ mechanisms where $\varepsilon ' = \varepsilon \sqrt{2k \log (1/\delta ')} + k\varepsilon (exp(\varepsilon) - 1)$.

In principle, shuffling can be implemented via multi-party computation (MPC), mixnets, running the shuffler on secure hardware or via a trusted third party \cite{movahedi2015secure,cheu2019distributed,bittau2017prochlo,allen2019algorithmic}. 

\subsection{Selecting a Privacy Amplification Model}
We provide some high level proposals:

\noindent \textbf{Shuffler:} can be used to generate text that's fed into linear classifiers with high utility. For example, a mechanism that outputs a sentiment class based on private perturbed data can still yield high utility on user de-linked and shuffled data. 

\noindent \textbf{Sub-sampler:} For other use cases such as personalization which require some form of user linked data, a sub-sampler can be used instead of a shuffler. This will be more suitable if the data is reasonably uniform (without outliers).

\noindent \textbf{K-threshold:} with randomized sub-sampling can be used for cases where the underlying data follows a long tail distribution such as for annotating data in crowdsourcing or training generalized ML models with user data.

\section{Proposal 2: Improved Randomizers}\label{sec:privacy_mechanism}
The randomizer $\mathcal{R}$ is based on the $d_{\mathcal{X}}$ metric privacy mechanism described by \cite{wsdm_paper} on word embeddings where the distance between word vectors is represented as the Euclidean metric. Alg.~\ref{alg:madlib} presents an overview of that mechanism. 
A similar mechanism was also proposed by \cite{fernandes2018generalised}, however, the distance metric was the Earth mover distance. Similarly, \cite{icdm_paper} extended the model to demonstrate preserving privacy using noise sampled from Hyperbolic space. The metric space of interest is as defined by word embedding models which organize discrete words in a continuous space such that the similarity in the space reflects their semantic affinity. Models such as \textsc{Word2Vec} \cite{mikolov2013distributed}, \textsc{GloVe} \cite{pennington2014glove}, and \textsc{fastText} \cite{bojanowski2016enriching} create such a mapping $\phi : \mathcal{W} \to \mathbb{R}^d$, where the distance function is expressed as $d : \mathcal{W} \times \mathcal{W} \to [0,\infty)$. The distance $d(w, w')$ between a pair of words is therefore given as $\norm{\phi(w) - \phi(w')}$, where $\norm{\cdot}$ is the Euclidean norm on $\mathbb{R}^d$.
  
%The privacy mechanism $\mathcal{R}$ of \cite{wsdm_paper} (see Alg~\ref{alg:madlib}) works by computing the vector representation $\phi(w)$ of a word $w$ in the embedding space, applying noise $N$ calibrated to the global metric sensitivity $\Delta_{\mathcal{M}_f}$ to obtain a \emph{perturbed} vector $\hat{\phi} = \phi(w) + N$, and then swapping the word $w$ with the word $\hat{w}$ whose embedding is closest to $\hat{\phi}$. The noise is calibrated over the entire size vocabulary $|\mathcal{W}|$ such that there is always a non zero probability of transforming one word into \emph{any} other word. The probability of the output distribution is scaled by the distance between the respective input words with larger probability mass concentrated closer to the word of interest. 

\begin{algorithm}
\footnotesize
\DontPrintSemicolon
\SetKw{KwRelease}{release}
\caption{Privacy Mechanism of \cite{wsdm_paper}}\label{alg:madlib}
\KwIn{string $x = w_1 w_2 \cdots w_{\ell}$, privacy param $\varepsilon > 0$}
\For{$i \in \{1, \ldots, \ell\}$}{
Word embedding $\phi_i = \phi(w_i)$\;
Sample noise $N$ with density $p_N(z) \propto \exp(- \varepsilon \norm{z})$.
%\sey(thanks, fixed) \zekun{is it $\exp(-\epsilon\|z\|)$? if so, also need to revise in section 5.2} \;
Perturb embedding with noise $\hat{\phi}_i = \phi_i + N$.\;% with noise density $p_N(z) \propto \exp(- \varepsilon \norm{z})$\;
Discretization $\hat{w}_i = \text{argmin}_{u \in \cW} \norm{\phi(u) - \hat{\phi}_i}$\;
Insert $\hat{w}_i$ in $i$th position of $\hat{x}$.\;
}
\KwRelease{$\hat{x}$}
\end{algorithm}

This mechanism however leads to sub-optimal accuracies due to a lack of uniformity in the embedding space. In particular, to achieve a certain level of privacy protection% in terms of proxy statistics
, the amount of noise is controlled by the worst-case word, which roughly corresponds to the word whose embedding is farther apart from any other word (\ie{} the global sensitivity). Therefore, at a given level of $\varepsilon$, a unique word like \emph{nudiustertian} will be perturbed similarly to a common word like \emph{drunk} which has over $2,000$ possible synonyms\footnote{\url{https://www.mhpbooks.com/books/drunk/}}.
To improve on this, we propose a variation of the original mechanism that can provide a fixed level of plausible deniability \cite{bindschaedler2017plausible}, measured in terms of the proxy statistics of \cite{wsdm_paper} with less noise, thus yielding more accuracy. In other words, the improved mechanisms should provide the same level of plausible deniability as the original mechanism, but under a larger value of $\varepsilon$. To achieve this goal, we propose three different strategies:
\begin{enumerate}
%\item Modify the word embeddings to have uniform density.
\item Defining a prior over the word embeddings to account for the space variability.
\item Calibrating the injected noise to the local sensitivity of the metric space. 
%\sey{reworded and hopefully clearer in the text below} \abhi{How are 2 and 3 different?} 
\item Adopting a truncated noise mechanism within an admissible region.
\end{enumerate}

\subsection{Density-Modulated Noise}
We observe that Alg.~\ref{alg:madlib} can be interpreted as an instance of the exponential mechanism \cite{mcsherry2007mechanism} together with a post-processing step (Alg.~\ref{alg:madlib}: Line $5$). Further, noise sampling via the exponential mechanism assumes a base measure $\mu(z)$ with a uniform distribution over the feasible range. Accordingly, line $3$ of Alg.~\ref{alg:madlib} can be expanded as  $p_N(z) \propto \mu(z) \times \exp(- \varepsilon \norm{z})$. However, the distribution of words in $\mathbb{R}^d$ is not uniform over the embedding space. As a consequence of Zipf's law, some words occur more frequently in a dataset and are surrounded by dense regions of similar words in the embedding space~\cite{gabaix1999zipf}. 

A natural way to ``bias" an exponential mechanism without changing its privacy properties is to modulate it with a public ``prior" $\mu(z)$. For example, such a prior can be constructed over a publicly available corpus such as Wikipedia or Common Crawl. 
%\abhi{What does public mean here? -- the attacker knows this prior?}. 
The question we address in this section is whether we can design an appropriate, potentially unnormalized, prior such that the resulting exponential mechanism that samples from $p_N(z) \propto \mu(z) \times \exp(- \varepsilon \norm{z})$ provides more accurate answers than the original mechanism under similar privacy constraints. An important research challenge in this direction is that by incorporating this correction to improve accuracy, we might end up with a mechanism that is computationally hard to sample from.

%% http://www.cse.psu.edu/~ads22/privacy598/papers/mt08.pdf
To obtain a prior that will solve the non-uniformity in the privacy mechanism using a vanilla word embedding is to modulate the distribution by a prior that captures the distribution of words in $\R^d$ induced by the word embedding. By introducing a prior that assigns high probability to dense areas of the embedding and low probability to sparse areas of the embedding, we can achieve the same level of plausible deniability statistics with smaller values of $\varepsilon$, hence, mitigating the worst-case effect that is observed in the unmodulated mechanism around sparse areas of the embedding.

One way to produce this prior measure $\mu(z)$ is to take a kernel density estimator with Radial Basis Function kernels on the resulting embedding, \ie{} $\mu(z) \propto \sum_{u \in \mathcal{W}} \exp\left( -\norm{z - \phi(u)}^2 / 2\sigma^2 \right)$ for some tuned variance $\sigma^2$.
%\sey{proposed} \abhi{Does this need a citation, or is this something we propose?}. 
However, it is not immediately clear how to sample from the modulated mechanism that on input $w$ has
density $p_N(z) \propto \mu(z) \times \exp(- \varepsilon \norm{z})$ for $\mu(z)$ defined above. Rather than sampling directly, we can either opt for an approximation to the distribution, or adopt indirect sampling strategies such as the Metropolis–Hastings algorithm \cite{metropolis1953equation,hastings1970monte}.

%Another method of compensating for the non-uniformity in the context of $d_{\mathcal{X}}$ privacy was introduced by \cite{chatzikokolakis2015constructing}. They use an elastic distinguishability metric that warps the geometrical distance (in $2-$dimensional Euclidean space), by capturing the density of each area. They then proceed to apply less noise to highly populated regions, and more noise to sparse regions as defined by the number of, and different types of nearby points .
%

Another observation is that we don’t need to pay the cost of an expensive sampling every time we want to use the mechanism. Instead, by introducing the projection step of the sampled vector to the closest word embedding, we can represent the mechanism by a $W \times W$ matrix containing the probabilities $\Pr[M(w) = w']$, where $M$ is the complete mechanism. We can precompute and store these $||W||^2$ probabilities and then use this matrix to define the output distribution every time we run the mechanism.

However, even though we can potentially make accuracy gains by incorporating the prior $\mu$ that captures the distribution of words in $\mathbb{R}^d$, we need to assume that the attacker is an informed adversary \cite{dwork2006calibrating}. Consequently, given that the adversary knows the prior (\eg{} since the word embeddings are public), the attacker's objective is to determine the user's actual word $w$, given the output $w'$ of the mechanism $\mathcal{R}$. The probability that $w$ is the word that generated $w'$ is given by the posterior probability distribution:

\begin{align*}
\Pr(w|w') = \frac{\mu(x) \cdot \Pr[\mathcal{R}(w) = w']}{\sum_{w \in \mathcal{W}}\mu(x) \cdot \Pr[\mathcal{R}(w) = w']}.
\end{align*}

%https://pdfs.semanticscholar.org/8c39/8b9e760faafc45c78122ae23d07c60985ea0.pdf
\noindent The privacy objective is to protect against a Bayesian adversary that can perform an optimal inference attack \cite{shokri2012protecting} of the form: 

\begin{align*}
\hat{w} = \argmin_{\hat{w} \in \mathcal{W}} \sum_{w \in \mathcal{W}} \Pr(w|w') d(\phi(\hat{w}), \phi(w))
\end{align*}

\noindent where word $\hat{w}$ is the adversary's inference given the output $w'$ of the randomizer, when the original word is $w$, and $d$ is the Euclidean distance between the word vectors. 
%\sey{not quite sure} \abhi{Interesting. Do we know of any works or have ideas on how to handle this adversary? Also, why delete the text after this? I think it makes sense to add it here.}

% BEGIN DELETED

%Therefore, compared to the original mechanism presented by \cite{wsdm_paper}, our updated randomizer introduces two main changes corresponding to the following lines in  Alg.~\ref{alg:madlib}: 
%\begin{enumerate}
%\item \textsc{Line $3$}: we sample the noise based on a \emph{restricted} local metric sensitivity measure over the vector space
%\item \textsc{Line $5$}: we utilize \emph{approximate} pairwise Euclidean distances between word vectors to provide stronger guarantees
%\end{enumerate}
%Our privacy model begins by creating an approximate preservation of the distance between words. Therefore, the $i$th nearest word to a given word $w$ becomes non deterministic, and the post processing step (line $5$ of \Cref{alg:madlib}) becomes stochastic for a given magnitude of noise. This provides additional privacy guarantees against the attack in \Cref{eqn:attack}. To achieve this, we use random projections to reconstruct the metric space of the vectors such that the pairwise distances between word representations are only approximately retained.

% END DELETED

\subsection{Calibrating Noise to Data Sensitivity}\label{sec:sensitivity}
%Consider two possible word vectors $\phi(w)$ and $\phi(w')$ in the embedding space equipped with a distance function $d$. We say that a randomizer $\mathcal{R} :\cW \to \cW$ satisfies $\varepsilon d_{\chi}$-privacy if for any $w, w' \in \cW$, the guarantees of Defn~\ref{defn:dp} are satisfied.

% the distributions over outputs of $\mathcal{R}(w)$ and $\mathcal{R}(w')$ satisfy the following bound: for all $w \in \cW$ we have
% \begin{align}\label{eqn:mdp}
% \frac{\Pr[\mathcal{R}(w) = \hat{w}]}{\Pr[\mathcal{R}(w') = \hat{w}]} \leq e^{\varepsilon d(w,w')} \enspace.
% \end{align}
In the proof of \cite{wsdm_paper}, $\hat{w}$ is calibrated at the worst-case distance $T$ from $w$ and $w'$ which is analogous to the global sensitivity. We can however, have a data dependent sensitivity definition over the metric space:

\begin{definition}[Local sensitivity \cite{nissim2007smooth}]\label{def:local_sensitivity}
The local sensitivity of a function $f : \mathcal{X}^n \to \mathbb{R}^d$ is given for $x \sim x' \in \mathcal{X}$ as,
\begin{align*}
\Delta_{\mathcal{L}_f} = \max\limits_{x' : d(x, x') = 1} \norm{f(x) - f(x')}_1
\end{align*}
\end{definition}

\noindent The local sensitivity of $f$ with respect to $x$ is how much $f(x')$ can differ from $f(x)$ for any $x'$ adjacent to the input $x$ (and not any possible entry $x$). We observe that the global sensitivity $\Delta_{\mathcal{G}_f} = \max_x \Delta_{\mathcal{L}_f}(x)$. However, a mechanism that adds noise scaled to the local sensitivity does not preserve \ac{DP} as the noise magnitude can leak information \cite{nissim2007smooth}. To address this, for example, \cite{nissim2007smooth} adds noise calibrated to a \emph{smooth bound} on the local sensitivity. The noise is typically sampled from the Laplace distribution. Thus, if we consider $\hat{w}$ at a distance $0 < t < T$, then the local sensitivity $\Delta_{\mathcal{L}_f}$ is:
\begin{align}\label{eqn:loc_sens}
{\Delta_{\mathcal{L}_f}}^{(t)} = \max\limits_{w' : d(w, w') \leq t} \Delta_{\mathcal{L}_f}.
\end{align} 
However, for our rare word  $w=$\emph{nudiustertian}, the local sensitivity might still leak information on output $\hat{w}$. 
As a result, we can construct the smooth sensitivity $\Delta_{\mathcal{S}_f}$ %\sey{updated} \abhi{undefined} 
as a $\beta-$smooth upper bound \cite{nissim2007smooth} on the local sensitivity. The desired properties of the bound include that: 

(1) $\forall w \in \cW : \hfill \Delta_{\mathcal{S}_f}(w) \geq \Delta_{\mathcal{L}_f}(w)$ 

(2) $\forall w,w' \in \cW : \hfill \Delta_{\mathcal{S}_f}(w) \leq e^\beta \cdot \Delta_{\mathcal{S}_f}(w')$

Observe that the smooth bound is equal to the global sensitivity $\Delta_{\mathcal{G}_f}$ when $\beta = 0$. Therefore, the smallest function $\Delta_{\mathcal{S}^{*}_{f,\beta}}$ that satisfies the two stated properties is the smooth sensitivity of the underlying function $f$ and can be stated as:
\begin{align*}
\Delta_{\mathcal{S}^{*}_{f,\beta}}(w) = \max\limits_{w' : d(w, w') \leq t} \big(\Delta_{\mathcal{L}_f}(w') \cdot e^{-\beta d(w, w')} \big)
\end{align*} 

%\subsection{Defining the local sensitivity}
%The local sensitivity ${\Delta_{\mathcal{L}_f}}^{(t)}$ can be replaced with another function $C^{(t)}(w)$ s.t. $\forall w,w' \in \cW : C^{(t)}(w) \leq C^{(t + \delta)}(w')$ as described by \cite{kasiviswanathan2013analyzing}. We obtain ${\Delta_{\mathcal{L}_f}}^{(t)}$ by calculating the average distance between the current word $w$, and its $k$ closest neighbors (rather than the max distance for $\Delta_{\mathcal{G}_f}$):
%\begin{align}
%t = \frac{1}{k} \sum_{i=1}^k \norm{\phi(w) - \phi(w_i)}
%\end{align} 

%\sey{I have tried to update it, let me know if I should take another pass} \abhi {the paragraph above is hard to read.} 

However, we cannot describe the local and smooth sensitivity this way since the local sensitivity construction in Def~\ref{def:local_sensitivity} was defined for integer-valued metrics (such as the Hamming distance). To translate this to real-valued metrics as is required for $d_{\chi}$-privacy, we can adopt the approach of \cite{laud2020framework} for defining the local sensitivity in metric spaces. 

First, we consider each word embedding vector as a point in some Banach space. A Banach space is a vector space with a metric that allows the computation of vector length and distance between vectors. For example, our $n-$dimensional Euclidean space of word embeddings, with the Euclidean norm is a Banach space. 
%\sey{yes} \abhi{Is this incomplete?}

Next, we observe from \cite{kasiviswanathan2013analyzing} that the local sensitivity of a function is similar to its derivative (\eg{} taking the limits in Eqn~\ref{eqn:loc_sens} as $t \to 0$). Therefore, the aim is to find an analog of a suitable derivative for continuous functions. One option is for the local sensitivity to be defined as the \emph{Fr\'echet derivative} in Banach spaces. \cite{laud2020framework} described an approach for this and they demonstrated how to apply noise sampled from the Cauchy distribution to satisfy the DP guarantees. Additional research would be needed to explore the direct application of the method of local (and then smooth) sensitivity calibration to embedding spaces.
\vspace{-1em}
\paragraph{Truncated Noise Mechanisms}
The standard $d_{\chi}$-privacy mechanisms were designed by
borrowing ideas from the privacy methods used for location data \cite{andres2013geo}. One of the proposed approaches in that work was to truncate the mechanism to report only points
within the limits of the area of interest. To achieve this, they define an `acceptable area' of admissible points $\mathcal{A} \subset \R^2$ (i.e., location privacy in $2-d$ space) beyond which results are truncated to the closest point in $\mathcal{A}$. Other truncation mechanisms have been explored in traditional DP including the truncated Laplacian \cite{geng2018truncated}, and truncated geometric mechanism within the context of $d_{\chi}$-privacy \cite{chatzikokolakis2013broadening}. 

Designing a corollary for text based $d_{\chi}$-privacy requires an approach to setting the truncation bounds while maintaining the privacy guarantees. We identify $2$ potential ways of achieving this: (1) Distance based truncation; and (2) K-nearest neighbor based truncation. To achieve (1) we can define a distance based limit similar to \cite{andres2013geo}. In this approach, a word can only get perturbed to words within the distance-defined admissible area $\mathcal{A} \in \mathcal{U}$. The maximum distance $\tau$ between a word and the farthest word in $\mathcal{A}$ is defined and fixed a-priori. To handle words that fall outside the noise limits, \cite{andres2013geo} proposes a discretization step to select the closest word in $\mathcal{A}$. Another option is for the mechanism to concentrate the probability of selecting a word to the admissible area $\mathcal{A}$, while assigning a residual probability to satisfy the \ac{DP} guarantee on the entire set $\mathcal{U}$. Therefore, when the noise exceeds the distance $\tau$, a replacement is randomly drawn from the set $\mathcal{U} - \mathcal{A}$. 

The downside of this approach is seen in regions of sharply varying density e.g., in the embedding space where one word has $2,000$ synonyms which potentially fall within $\mathcal{A}$ and the rare word with no neighbors in $\mathcal{A}$. Therefore to achieve (2) rather than having a fixed distance from each word, we can also define the (randomized) $k-$closest words as delineating our acceptable area. One potential benefit of this is, in dense spaces, we can select closer candidates while simultaneously guaranteeing that isolated words are replaced by one of their $k-$nearest neighbors regardless of how far off it is.

Implementing either of these mechanisms however come with their own set of challenges. For example, there isn't a direct way to set a maximum distance when drawing the multivariate laplacian noise that was proposed by \cite{wsdm_paper}. One option will be to fix $\tau$, or the distance to the max randomized $k$ as the local sensitivity described in Sec~\ref{sec:sensitivity}. Another option will be to rethink the entire design of the randomizers such that the noise is not added to the vector representation of the words, but to these $\tau$ distances.
\vspace{-1em}
\paragraph{Connections to Related Work}
The traditional \ac{DP} literature contains techniques to limit the privacy preserving noise added to a mechanism. Just as with the median example, these approaches take into consideration the actual dataset as opposed to the worst case guarantees of a possible theoretical construct. In one work, \cite{nissim2007smooth} introduced the notion of smooth sensitivity where a smooth upper bound on the local sensitivity is used to determine how much noise is added. They demonstrated this method for privately computing the median of a dataset, and the number of triangles in a graph. %The technique however breaks down when the dataset contains exponentially far outliers (an occurrence which is rather suited to standard \ac{DP} mechanisms).

Similarly, \cite{dwork2009differential} introduced a paradigm called \emph{Propose-Test-Release} (PTR) where: the algorithm proposes a bound on sensitivity, tests the adequacy of the bound on the given dataset, and halts if the sensitivity is too high. The PTR technique is connected to \emph{robust statistics} which is concerned with outliers and rounding errors in a dataset \cite{dwork2010differential}. %As a corollary to \ac{DP}, constructing the sensitivity around robust statistics serves to ensure that the private data of an individual does not significantly change the distribution of the outputs on the analysis. This technique also fails when the typical sensitivity is high such as when computing sums or averages \cite{amin2019bounding}.

In related work, \cite{kasiviswanathan2013analyzing} extended the notion of limiting the noise for private graph analysis where the degree bound (a function of the number of nodes in the graph) can be arbitrary. To achieve this, they set a $D$ bound on the graph which aims to keep the sensitivity low while retaining as large a fraction of the graph as possible. Similarly within the context of graphs (specifically, social networks), \cite{blocki2013differentially} introduce the idea of \emph{restricted sensitivity} without setting a bound on the degree of the graph. Instead, they define a hypothesis (a subset of all possible datasets) over which they compute the sensitivity.

These all describe principled approaches to limit the magnitude of noise applied to a privacy preserving mechanism in the contexts of statistical and graph analysis by redefining the sensitivity that controls the noise. %In this work, we are concerned with limiting noise within privacy mechanisms that operate over vectorial representations of words. 
As opposed to the reviewed techniques, our representations are within a metric space defined by word embeddings. The sensitivity is therefore described over the vocabulary of the embeddings. Therefore, unlike statistical analysis where releasing noisy counts still contains informative value, adding noise to word representations, can result in a rapid degradation of utility. 

%This is because, to preserve privacy, there has to be a non-negligible probability of one word being transformed into any other, no matter how unrelated they are \cite{wsdm_paper}. 

%\abhi{This section reads more like discussion than related work.}
%\sey{updated and shortened, can have a second look}

\section{Conclusion and Future work}
In this proposal paper, we surveyed some of the challenges of building differentially private mechanisms for generating text based on word embeddings. We investigated approaches built on the $d_{\chi}$-privacy framework in Euclidean space. The core issues stem from the non-uniformity of the metric space defined by embeddings and the need to provide worst case guarantees for outliers as required by differential privacy. This necessitates a large amount of noise thus leading to utility impacts on downstream tasks that rely on the generated text as input features.

Our approach was to explore the resulting utility issues from different perspectives: first, considering methods of reducing the required noise by deferring additional guarantees to other privacy amplification mechanisms that do not require noise (such as shuffling). We then proposed three ways to reduce the needed noise by accounting for the density around the word under consideration. These included introducing a prior, re-calibrating the noise, or truncating the noise. In future work, we plan to explore these approaches in detail and provide a study on what works, when it works, and why. Our aim is to provide a principled approach to studying these mechanisms in order to accelerate the research and drive adoption.

%In this paper, we presented an approach to carry out privacy preserving machine learning on models that require text based data. As compared to traditional metric \ac{DP} algorithms, the noise in our mechanism is sampled from the smooth sensitivity defined in the embedding space. The result of this yields noise modulated to the density of the embedding space, therefore improving utility by side-stepping worst-case guarantees.

%In future work, we plan to improve the method by adding a dimensionality reduction step on the word embeddings \abhi{Why not add a discussion to the paper now? Maybe you can borrow some text from your paper with Shiva on JL transformations?}. This reduces the $n \times d$ dimensional 
%embeddings to an $n \times k$ dimensional vector space. Reducing the dimensionality of the embeddings will lead to less noise, especially at smaller values of $\varepsilon$. We will show the tradeoff with the new dimensionality as we expect that at a certain point, the reduction will lead to increased utility loss.

\bibliographystyle{alpha}
\bibliography{ref}

\end{document}